\RequirePackage{fix-cm}
\documentclass[smallextended]{svjour3}       
\smartqed  
\usepackage{graphicx}

\usepackage{amsmath}
\usepackage{graphicx,psfrag,epsf}
\usepackage{enumerate}
\usepackage{natbib}
\usepackage{url} 

\usepackage{xcolor}

\usepackage{hyperref}
\usepackage{times}
\usepackage{bm}
\usepackage{natbib}

\usepackage{graphicx}
\usepackage{url}
\usepackage{amsmath}
\usepackage{amsfonts}
\usepackage{amssymb}
\usepackage{float}
\usepackage[plain,noend]{algorithm2e}

\makeatletter
\newcommand*{\rom}[1]{\expandafter\@slowromancap\romannumeral #1@}
\makeatother

\begin{document}

\title{Estimating the Number of Clusters via Normalized Cluster Instability
}

\titlerunning{Normalized Cluster Instability}        

\author{Jonas M. B. Haslbeck         \and
        Dirk U. Wulff 
}


\institute{Jonas Haslbeck \at
              University of Amsterdam \\
              Psychological Methods \\
              Nieuwe Achtergracht 129-B \\
              Postbus 15906 \\
              1018 WT, Amsterdam \\
              the Netherlands \\
              \email{jonashaslbeck@gmail.com} \\
             \url{http://jmbh.github.io} \\
           \and
           Dirk Wulff \at
           Center for Cognitive and Decision Science\\
           University of Basel\\
           \email{dirk.wulff@gmail.com}
}

\date{Received: date / Accepted: date}

\maketitle

\begin{abstract}
We improve current instability-based methods for the selection of the number of clusters $k$ in cluster analysis by developing a normalized cluster instability measure that corrects for the distribution of cluster sizes, a previously unaccounted driver of cluster instability. We show that our normalized instability measure outperforms current instability-based measures across the whole sequence of possible $k$ and especially overcomes limitations in the context of large $k$. We also compare, for the first time, model-based and model-free approaches to determine cluster-instability and find their performance to be comparable. We make our method available in the R-package \verb+cstab+.

\end{abstract}

\section{Introduction}

A central problem in cluster analysis is selecting the appropriate number of clusters $k$. To develop a method that identifies the true number of clusters $k^*$, one requires a formal definition for what a 'good clustering' is. Different definitions have been proposed and it is generally accepted that the appropriate definition depends on the clustering problem at hand \cite[see e.g.][]{friedman2001elements, hennig2015true}.

Most definitions in the literature define the quality of a clustering in terms of a distance metric between the clustered objects. These methods select $k$ by trading-off a function of this distance metric against the number of $k$. The most commonly used distance metric is the average (across clusters) within-cluster dissimilarity $W(k)$, which is the average dissimilarity between all object-pairs within the same cluster. When selecting the $k$ based on this metric it is assumed that $W(k)$ would exhibit a 'kink' at $k = k^*$, i.e., when the assumed number of clusters $k$ matches the true number of clusters $k^*$. This is because for $k = k^*$ clusters are highly homogenous allowing for only minimal improvements in $W(k)$ when splitting objects further. All methods in this class, in one way or another, aim to identify this 'kink'. Two examples are the Gap statistic \citep{tibshirani2001estimating} and the Jump statistic \citep{sugarfinding}. Related metrics are the Silhouette statistic \citep{rousseeuw1987silhouettes}, which is a measure of cluster separation rather than variance, and a refinement thereof, the Slope statistic \citep{fujita2014non}.

Another approach, which is the focus of this paper, defines a good clustering in terms of its \textit{stability} against small pertubations of the data. Correspondingly, stability-based methods select the $k$ yielding the most stable clustering. Stability-based methods are attractive because they do not require a metric for the distance between objects and have been shown to perform at least as well as state-of-the-art distance-based methods \citep{ben2001stability, tibshirani2005cluster, hennig2007cluster, wang2010consistent, FangWang2012}. 

In the present paper, we show that two previously proposed stability-based approaches, the \emph{model-based} approach \citep{FangWang2012} and the \emph{model-free} approach \citep{ben2001stability}, depend heavily on the distribution of cluster sizes $M$ leading to incorrect estimates $\hat{k}$ especially when the list of candidate $k$ is not restricted to small numbers. To address this problem, we develop a normalized cluster instability measure correcting for the influence of $M$. We show that our normalized instability measure outperforms current instability-based measures across the whole sequence of possible $k$. We also compare, for the first time, model-based and model-free approaches to determine cluster-instability and find their performance to be comparable. We make our method available in the R-package \verb+cstab+, which is available on The Comprehensive R Archive Network (CRAN).

\section{Clustering Instability}

A clustering $\psi(\cdot)$ is stable when it is robust against perturbations of the data. Stable clusterings, thus, assign two objects $X_1, X_2$ that occupy the same cluster in the clustering $\psi_a(X)$ based on the original data $X$ to be in same cluster in a clustering $\psi_b(\widetilde{X})$ based on a perturbed data $\widetilde{X}$ and vice versa for objects not occupying the same cluster. In this section, we formalize this idea following \cite{wang2010consistent} using based on \textit{clustering distance} and \textit{clustering instability}.

Let $\mathbf{X} = \{X_1, \dots, X_n \} \in \mathbb{R}^{n \times p}$ be $n$ samples from an unknown distribution $\mathcal{P}$ defined on $\mathbb{R}^p$. We define a clustering $\psi: \mathbb{R}^{n \times p} \mapsto \{1, \dots, K\}^n$ as a mapping from a configuration $X_i \in \mathbb{R}^p$ to a cluster assignment $k \in \{1, \dots, K\}$. A clustering algorithm $\Psi(\mathbf{X}, k)$ learns such a mapping $\psi$ from data.

\begin{definition}[Clustering Distance]\label{def:distance}
	The distance between any pair of clusterings $\psi_a(X)$ and $\psi_b(X)$ is defined as
	
	$$
	d(\psi_a(X^0), \psi_b(X^0))  = 
	| \mathbb{I}_{\{ \psi_a(X_1) = \psi_a(X_2) \}} - \mathbb{I}_{\{ \psi_b(X_1) = \psi_b(X_2) \}} | 
	,
	$$
	
	\noindent
	where $X^0 = \{ X_1, X_2 \}$ is a fixed vector containing the two objects $X_1, X_2 \in \mathbb{R}^p$  and $\mathbb{I}_{ \{E \} }$ is the indicator function for the event $E$.
	
\end{definition}

\begin{figure}[h]
	\centering
	\includegraphics[width=.7\linewidth]{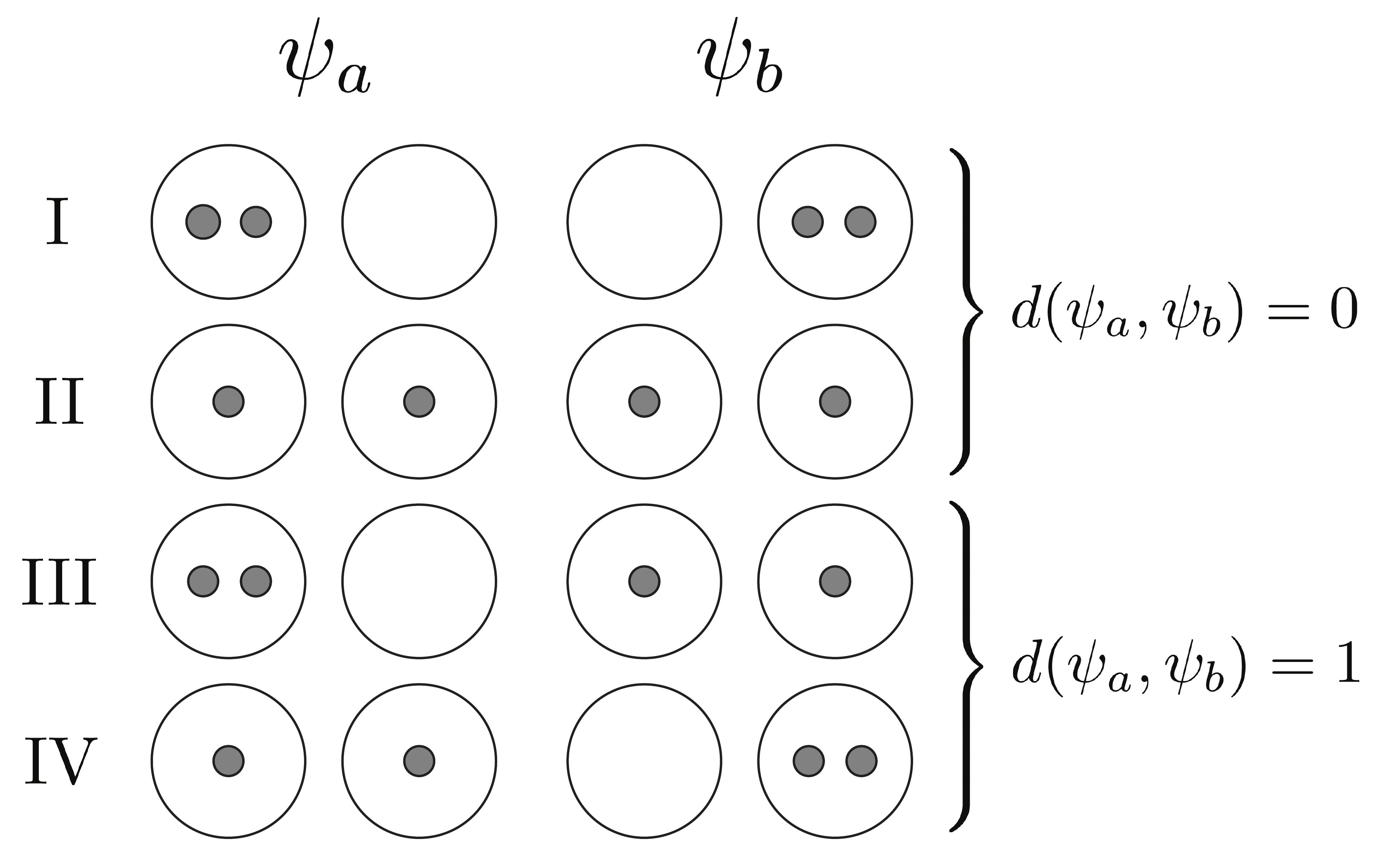}
	\caption{The clustering distance for the four possible configurations of clustering assignments of clusterings $\psi_a, \psi_b$ for two objects $X_1, X_2$.}
	\label{fig:distance_cases}
\end{figure}

\noindent
Figure \ref{fig:distance_cases} illustrates the four possible cases that contribute to the clustering distance. To the extent that $\psi_a, \psi_b$ agree on whether any two objects $X_1, X_2$ occupy the same cluster or not (\rom{1} and \rom{2}), the distance approaches zero. Conversely, to the extent that $\psi_a, \psi_b$ disagree (\rom{3} and \rom{4}), the distance approaches 1. The clustering distance, thus, reflects the relative proportion of cases \rom{1} and \rom{2} versus cases \rom{3} and \rom{4}. Accordingly, clustering distance can also be expressed as:

\begin{equation}\label{def:distance2}
\begin{split}
d(\psi_a, \psi_b)  = 
&\; P \left [ \psi_a(X_1) = \psi_a(X_2), \psi_b(X_1) \neq \psi_b(X_2)   \right ]\\  
+ & \;P \left [ \psi_a(X_1) \neq \psi_a(X_2), \psi_b(X_1) = \psi_b(X_2)   \right ].
\end{split}
\end{equation}

\begin{definition}[Clustering instability]\label{def:instability}
	We define the clustering instability of clustering algorithm $\Psi(X, k )$ as
	
	$$
	s(\Psi, \mathbf{X}, k)  = \mathbb{E}_\mathcal{P}  \left [  
	\frac{1}{n(n-1)/2}
	\sum_{i,j = 1, i \neq j}^{n} 
	d (\psi_a(X_i), \psi_b(X_j) ) 
	\right ]
	,
	$$
	
	\noindent
	where $\mathbf{X} \in \mathbb{R}^{n \times p}$ is the original data drawn from an unknown distribution $\mathcal{P}$ defined on $\mathbb{R}^p$. The clusterings $\psi_a$ and $\psi_b$ are obtained from two independent samples $\widetilde{\mathbf{X}}_a$ and $\widetilde{\mathbf{X}}_b$ drawn from $\mathcal{P}$. The expectation is taken with respect to $\mathcal{P}$.
\end{definition}

As the arithmetic over clustering distances, it holds that $s(\Psi, \mathbf{X}, k) \in [0,1]$. We estimate the true number of clusters $k^*$ by finding the minimum of the cluster instability $s(\Psi, \mathbf{X}, k) \in [0,1]$ across a sequence of $k$

\begin{equation}\label{eq:minimize}
\hat{k} = \arg \min_{2 \leq k \leq n}  s(\Psi, \mathbf{X}, k).
\end{equation}

Note that, when we defined clustering instability (\ref{def:instability}, we used two independent samples $\widetilde{\mathbf{X}}_a$, $\widetilde{\mathbf{X}}_b$ from the distribution $\mathcal{P}$ to obtain clusterings from "perturbed" data. Of course, in practice the distribution $\mathcal{P}$ is unknown. Following \cite{FangWang2012}, we resolve this issue using a bootstrap approach. Specifically, we take samples from the original data $\textbf{X}$ and treat them as surrogates for independent samples $\mathcal{P}$. 

The bootstrap samples will usually to contain different sets of objects - we estimate the expected\footnote{For large $n$, we have $P(a_j \in B_1 \land a_j \in B_2) = P(a_j \in B_1) P(a_j \in B_2) = (1 - \frac{1}{e})^2 \approx 0.400$} proportion of objects shared across bootstrap samples to be $\approx 0.4$. This leaves two routes for computing from them a clustering distance: to only rely on the objects shared across bootstraps, which we call the model-free approach, or to rely on model-based cluster predictions to derive cluster assignments for all objects across both bootstraps, which we call the model-based approach. The next sections describes these approaches in detail.

\section{Model-Based Clustering Instability}

The model-based approach to computing clustering instability uses clustering algorithms $\Psi(\cdot, k )$ to learn a partitioning of $\mathbb{R}^p$ into $k$ non-empty subsets (clusters) for each of the two perturbed data sets $\widetilde{\mathbf{X}}_1, \widetilde{\mathbf{X}}_2$. These partitions are then used to assign unseen objects to the clusters. An example for a model-based clustering algorithm is the k-means algorithm, which partitions $\mathbb{R}^p$ into $k$ Voroni cells \citep{hartigan1975clustering}. 

This approach was first described as a cross validation (CV) scheme \citep{tibshirani2005cluster, wang2010consistent}. Here we present the algorithm for the non-parametric bootstrap, which has been shown to perform better than CV \citep{FangWang2012}.\\

\begin{algorithm}[H]
	\caption{Model-Based Clustering Instability} \label{algorithm1}
	\vspace*{-12pt}
	\begin{enumerate}
		\item Take bootstrap samples $\widetilde{\mathbf{X}}_{a}, \widetilde{\mathbf{X}}_{b}$ from the empirical distribution $\mathbf{X}$
		\item Compute clustering assignments $\psi_a(\widetilde{\mathbf{X}}_{b}), \psi_b(\widetilde{\mathbf{X}}_{b})$ using the clustering algorithm $\Psi(\cdot, k )$
		\item Use the clusterings $\psi_a, \psi_b$ to compute assignments $\psi_a(\mathbf{X}), \psi_b(\mathbf{X})$ on the original data $\mathbf{X}$
		\item Use $\psi_a(\mathbf{X}), \psi_b(\mathbf{X})$ to compute the clustering instability $s_i$ as in (\ref{def:instability})
	\end{enumerate}
	Repeat 1-4 $B$ times and return the average instability $\bar{s}(\Psi, \mathbf{X}, k) = B^{-1} \sum_{i=1}^{B} s_i(\Psi, \mathbf{X}, k)$.
	\vspace*{12pt}
\end{algorithm}

\vspace{.5cm}

The model-based approach can also be used with spectral clustering \citep{ng2002spectral} using the method described in \citet{bengio2003spectral}. The major drawback of the model-based approach in Algorithm \ref{algorithm1} is that step 4 requires a complete partitioning of $\mathbb{R}^p$, rendering unavailable popular algorithms such as, for instance, hierarchical clustering \citep{friedman2001elements}. One could nonetheless employ the model-based procedure, in these cases, using an additional classifier (e.g. k nearest neighbors) to predict unseen objects. However, it may be more elegant to sidestep this issue using the model-free approach described below.

\section{Model-Free Clustering Instability}

The model-free approach \citep{ben2001stability} uses any clustering algorithm $\Psi(\cdot, k )$ to compute two clusterings on the two bootstrap samples $\widetilde{\mathbf{X}}_a, \widetilde{\mathbf{X}}_b$.
Clustering instability is then only computed for the intersection $\widetilde{\mathbf{X}}_{a \cap b} =  \widetilde{\mathbf{X}}_{a} \cap \widetilde{\mathbf{X}}_{b}$, that is, the set of objects that are shared across both bootstrap samples. This approach avoids the problem of having to assign unseen objects, because it only consider objects that are in both clustering. \\

\begin{algorithm}[H]
	\caption{Model-Free Clustering Instability} \label{algorithm2}
	\vspace*{-12pt}
	\begin{enumerate}
		\item Take bootstrap samples $\widetilde{\mathbf{X}}_{a}, \widetilde{\mathbf{X}}_{b}$ from the empirical distribution $\mathbf{X}$
		\item Compute clustering assignments $\psi_a (\widetilde{\mathbf{X}}_a), \psi_b(\widetilde{\mathbf{X}}_b)$ using the clustering algorithm $\Psi(\cdot, k )$
		\item Take the intersection $\widetilde{\mathbf{X}}_{a \cap b} =  \widetilde{\mathbf{X}}_{a} \cap \widetilde{\mathbf{X}}_{b}$
		\item Use $\psi_a(\widetilde{\mathbf{X}}_{a \cap b}), \psi_b(\widetilde{\mathbf{X}}_{a \cap b})$ to compute the clustering instability $s_i$ as in (\ref{def:instability})
	\end{enumerate}
	Repeat 1-4 $B$ times and return the average instability $\bar{s}(\Psi, \mathbf{X}, k) = B^{-1} \sum_{i = 1}^{B} s_i(\Psi, \mathbf{X}, k)$.
	\vspace*{12pt}
\end{algorithm}

\vspace{.5cm}

Because no unseen objects need to be assigned, Algorithm~\ref{algorithm2} can be inmplemented with \textit{any} clustering algorithm $\Psi(\cdot, k )$. A potential prize for this flexibility is that Algorithm~\ref{algorithm2} compared to Algorithm~\ref{algorithm1} computes clustering instability only on approximately 40 \% of the original data, implying that more bootstrap comparisons may be needed to achieve equal performance.

\section{Normalized Clustering Instability}

In this section, we introduce a normalized clustering instability (\ref{def:instability}) for both the model-based and the model-free approaches to remedy a shortcoming of current stability-based methods: clustering instability (as defined in \ref{def:instability}) has the undesirable property that it depends substantially on the distribution of cluster sizes $M$ and by extension the number of clusters $k$, irrespective of the true number of clusters $k^*$. To see this, consider the clustering distance in (\ref{def:distance}): for this distance to be nonzero two objects $X_1, X_2$ must be able to change their cluster assignment with regard to each other. This excludes $k=1$, where $X_1, X_2$ are forced to occupy the same cluster, and $k=n$, where $X_1, X_2$ are forced to occupy different clusters. Considering now the case of $k=n-1$, then the majority of objects will again be bound to occupy different clusters, expect for two that could occupy the same 2-object cluster in both clusterings or not. As a result, the clustering instability will, on average, be larger for $n-1$ than $n$, but not by much. Conversely, the only way for the clustering instability to be large is in the presence of moderately sized $k$ and relatively even cluster sizes. This means that in order to properly evaluate clustering instability one needs to take account of the distribution of cluster sizes $M$. Below we work out the exact relationship between $s(\Psi, \mathbf{X}, k)$ and $M$ in order to develop a new, normalized clustering instability measure. 

\begin{figure}[H]
	\centering
	\includegraphics[width=1\linewidth]{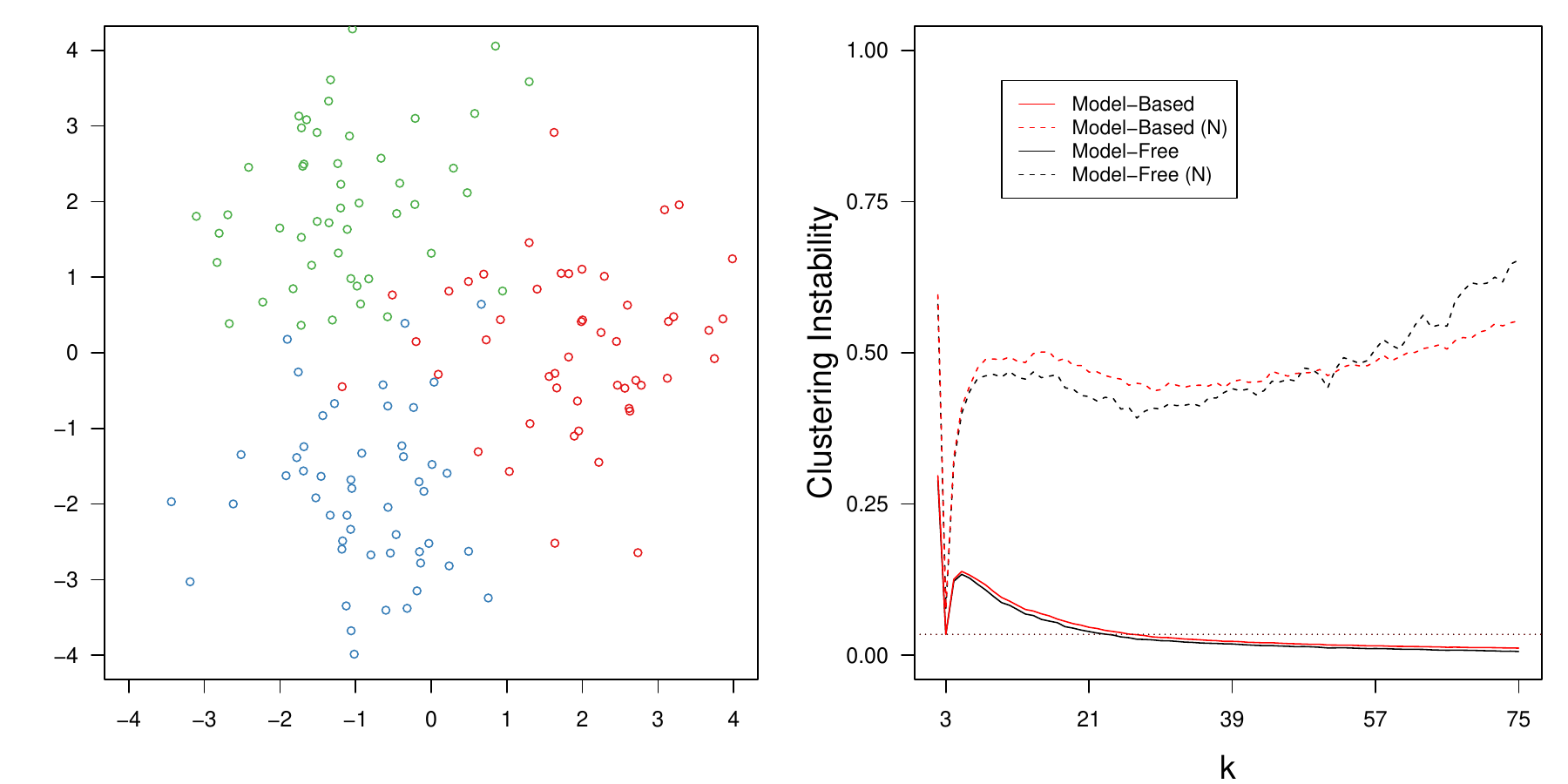}
	\caption{Left: mixture of three (each n = 50) 2-dimensional Gaussians with zero covariance and $\sigma_i = 1$. Right: instability path for the model-based (red) and model-free instability approach (black), both normalized (dashed) and unnormalized (solid). The horizontal lines indicate the local minimum of the instability path at $k^* = 3$ for each method. The estimate $\hat{k}$ will be incorrect (too large) if we consider $k$s with an instability below the corresponding horizontal line. } 
	\label{fig:illustration}
\end{figure}

First, however, we demonstrate possibly the most serious problem associated with the dependency of $s(\Psi, \mathbf{X}, k)$ on $M$. Figure \ref{fig:illustration} shows a typical instability path obtained for the clustering problem previously studied by \cite{FangWang2012}. The path shows a local minimum at $k = k^* = 3$, the true number of clusters, but beginning at $k = 6$ clustering instability begins to drop. Critically, instability crosses the value of instability for $k = 3$ (dashed lines) at $k = 23$ (model free) and $k = 25$ (model based). This is a general pattern of (unnormalized) cluster instability paths and a result of its dependency on $M$. As a consequence, $\hat{k}$ will substantially overestimate $k^*$, when the candidate set is not restricted to small $k$s. One approach to address this issue is to restrict the set of candidate $k$ to small numbers, as was done in previous publications, e.g., \cite{FangWang2012} and \cite{ben2001stability}. As it is difficult to know it difficult to know, however, when $s(\Psi, X, k)$ drops below $s(\Psi, X, k^*)$ -- the exact point of intersection will vary as a function of a variety of factors including the number of true clusters and cluster separation -- this approach will unlikely be very reliable. Importantly, this approach will take account of the complex relationship between $M$ and $s(\Psi, \mathbf{X}, k)$, impacting the latter across the entire range of $k$ and corrupting its signal of $k^*$. 

We now show how $d(\psi_a, \psi_b)$ as defined in (\ref{def:distance2}) is related to $M$ and by extension $k$, irrespective of the number of true clusters $k^*$. Let $M = \{n_1, n_2, \dots n_K \}$ be the number of objects in each of the clusters in a given clustering $\psi(\mathbf{X})$. Assuming independence between $\psi_a(\dot)$ and $\psi_b(\dot)$ given independently drawn $\widetilde{\mathbf{X}}_a$, $\widetilde{\mathbf{X}}_b$ we begin by rewriting the clustering distance (\ref{def:distance2}) as

\begin{equation}\label{def:distance3}
\begin{split}
d(\psi_a, \psi_b)  =
&\; P \left [ \psi_a(X_1) = \psi_a(X_2) \right ] *  P \left [ \psi_b(X_1) \neq \psi_b(X_2)   \right ] \\  
+ & \;P \left [ \psi_a(X_1) \neq \psi_a(X_2) \right ] *  P \left [ \psi_b(X_1) = \psi_b(X_2)   \right ]).
\end{split}
\end{equation}

\noindent
From (\ref{def:distance3}) we see that the clustering distance depends essentially on two probabilities, the probability that two objects $X_1, X_2$ are assigned to the same cluster $P \left [ \psi(X_1) = \psi(X_2) \right ] $ and its complement $P \left [ \psi(X_1) \neq \psi(X_2) \right ]$, which we can express using the former, i.e., $ P \left [ \psi(X_1) \neq \psi(X_2) \right ] = 1-P \left [ \psi(X_1) = \psi(X_2) \right ] $. Thus, we can express  (\ref{def:distance3}), using a single probability, the probability of two objects occupying the same cluster. In practice this probability will depend heavily on features of the data generating process such as e.g., cluster separation, making it possible to use (\ref{def:distance3}) as a signal of $k^*$. However, as we will work out below, $P \left [ \psi(X_1) = \psi(X_2) \right ]$ is also a function of $M$. 

Using combinatorics, we will express $P \left [ \psi(X_1) = \psi(X_2) \right ]$ as the ratio of the number of possible clusterings $N_{pair}$ for which $X_1, X_2$ are in the same cluster and the number of all possible clusterings $N_{tot}$, each of which depending exclusively on $M$, 

\begin{equation}\label{eq:prand}
P \left  [  \psi(X_1)  = \psi(X_2)   \right ]  = \frac{N_{pair}}{N_{tot}}.
\end{equation}

\noindent
We compute

\begin{equation}
N_{tot} = \prod_{1 \leq i \leq K }  {{m_i}\choose{n_i}},
\end{equation}

\noindent
where $n_i$ is the number of objects in cluster $i$ and $m_i$ is the number of objects that are not yet contained in already considered clusters\footnote{Note that in the model-based approach the sum $\sum_{i=1}^{K} n_i$ is equal to the number of objects in the original data set $\mathbf{X}$, whereas in the model-free approach it is equal to the number of objects in the intersection $\widetilde{\mathbf{X}}_{\alpha \cap \beta, b}$ of the two bootstrap samples.}

$$
m_i=\sum_{i \leq j \leq k }{n_j}.
$$

Intuitively, this counts all possible ways to select $n_1$ objects from the set of all objects, which then is multiplied by the number of possible ways to select $n_2$ objects from the set of all objects minus the just selected $n_1$ objects, etc.

Further, we compute $N_{pair}$

\begin{equation}
N_{pair}=\sum_{1 \leq i \leq k \atop n_i \geq 2}
{m_i-2\choose{n_i-2}}
\prod_{1 \leq j \leq k \atop j \neq i}
{m_j\choose{n_j}}.
\end{equation}

Here, the first term assumes two objects to occupy the same cluster, while the second term, analogous to above, computes the number of possible ways to distribute the remaining objects across clusters, respecting the cluster sizes $M$.

Together, $N_{tot}$ and $N_{pair}$ (\ref{eq:prand}) gives us $P \left [ \psi(X_1) = \psi(X_2) \right ] $, the probability that two objects $X_1, X_2$ are assigned to the same cluster by chance. Plugging this result into the definition of clustering distance in (\ref{def:distance3}) we obtain $d^r(\psi_a, \psi_b)$, the expected clustering distance under random object allocation for a given $M$ and $k$. 

\begin{figure}[h]
	\centering
	\includegraphics[width=.6\linewidth]{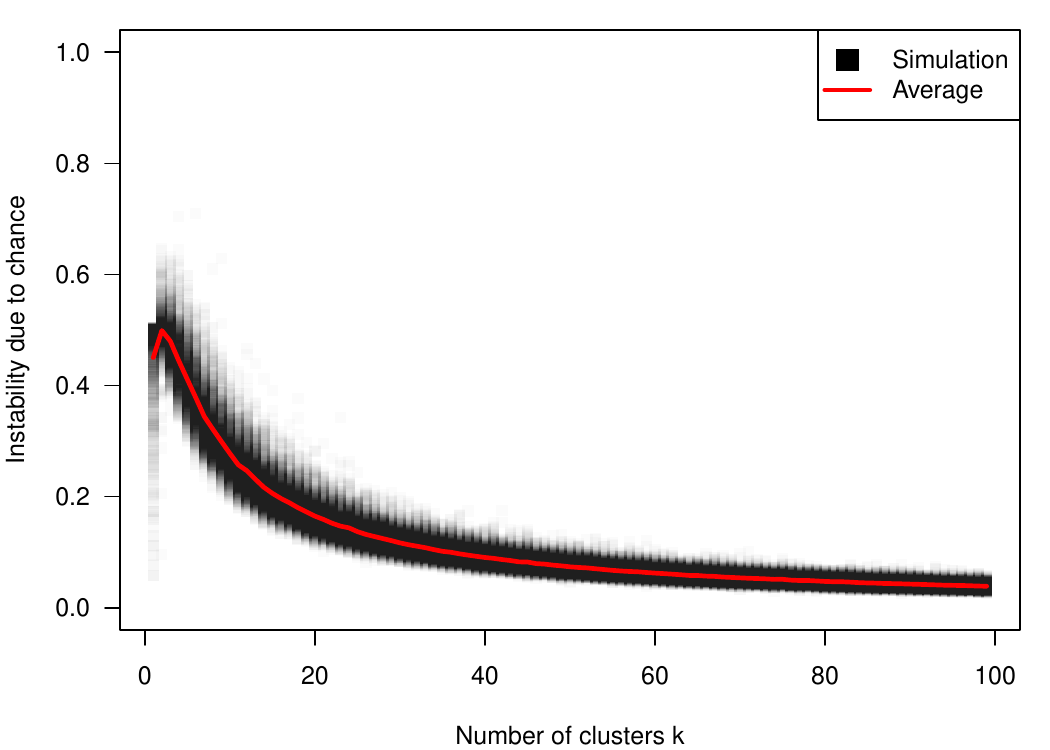}
	\caption{Instability due to chance for randomly generated $M$ for $k \in \{1, \dots, 100 \}$. Simulation is based on calculating instability for sets of $M$ generated by randomly distributing 100 objects across $k$ clusters. The shaded squares in the background show the distribution in $d^r$ only due to variation in $M$ for a given $k$.}
	\label{fig:instabcurve}
\end{figure}

Having defined $d^r(\psi_a, \psi_b)$, we can study its behaviour as a function of $M$ and $k$. Figure \ref{fig:instabcurve} shows $d^r$ for $k \in \{2, 3, \dots, 100\}$ and $M$ $\sim$ Multinomial($\boldsymbol{\theta}$) with $\boldsymbol{\theta}$ $\sim$ Dirichlet($\boldsymbol{1}$). Three key results emerge. First, $d^r$ is a non-monotonic function of $k$ showing maximal instability at $k=3$. Second, the largest change in $d^r$ occurs in the range of $k = [1, 10]$ and, thus, the range typically studied. This highlights the relevance of $d^r$ for small values of $k$. Third, $d^r$ decreases consistently to 0 as $k$ grows large. Note that $d^r$ essentially represents an upper bound for $d$, in the sense that non-random data with the same $M$ is expected to show $d < d^r$. Consequently, the tail behaviour of $d^r$ in \ref{fig:instabcurve} implies that $d$ will also approach 0 as $k$ grows large and, thus, eventually undercut any local minimum present for smaller $k$s as shown in Figure \ref{fig:illustration}.     

The above results demonstrate the importance of accounting for $d^r$ when inferring $k^*$ based on cluster instability. Following the interpretation of $d^r$ as the upper bound of $d$, we propose to correct for $d^r$ by computing a normalized clustering instability $d^n$ defined as

\begin{equation}
d^n(\psi_a, \psi_b) = \frac{d(\psi_a, \psi_b)}{d^r(\psi_a, \psi_b)}
.
\end{equation}

\noindent
The right panel of Figure \ref{fig:illustration} demonstrates how this new measure improves the identification of $k^*$. First, in comparison, to the unnormalized cluster instability, $d^n$ shows a much more distinct drop at $k=k^*=3$ rendering the result less susceptible to the influence of noise. Second, the path of the normalized cluster instability (dashed line) does not decrease for larger $k$ rendering $k=k^*=3$ a global optimum. Thus, normalized cluster instability $d^n$ facilitates consistent estimation of $k^*$ across the entire range of $k$. In the next section, we use numerical experiments to demonstrates the benefit of using normalized cluster instability $d^n$ in realistic settings. 

\section{Numerical experiments}

We now turn to numerical evaluation of the performance of unnormalized and normalized instability-based methods across four scenarios. This will include a comparison of the both instability-based approaches to the performance of four popular distance-based methods for selecting $k^*$.

\subsection{Data generation}\label{sim_data}

\begin{figure}[H]
	\centering
	\includegraphics[width=1\linewidth]{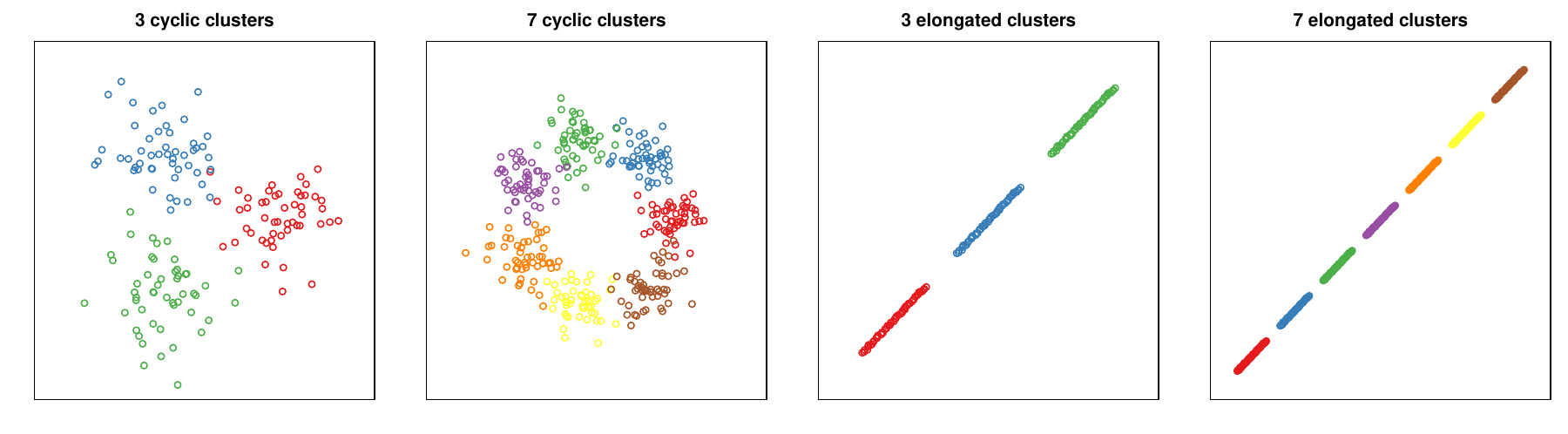}
	\caption{First column: three Gaussians with $\sigma = .1$ and $n=50$ placed on a circle; second column: seven Gaussians with $\sigma = .04$ and $n=50$ placed on a circle; third column: three elongated clusters in three dimensions (only the first 2 shown); fourth column: seven elongated clusters in three dimensions (only the first 2 shown).}
	\label{fig:simsetup}
\end{figure}

We generate data from Gaussian mixtures as illustrated in Figure \ref{fig:simsetup}. For the first scenario with $k^*=3$, we distributed the means of three Gaussians with $\sigma = .15$ arranged at equal distances on a unit circle and sampled $n=50$ from each Gaussian. The second scenario with $k^*=7$, we distributed the means of seven Gaussians with $\sigma = .04$ at equal distances on a unit circle and sampled $n=50$ from each Gaussian. This means that the total sample size of these two problems is 150 and 350, respectively. We chose prime numbers and a circular layout to avoid local minima for $k < k^*$. The third and fourth scenario used elongated clusters similar to those in \cite{tibshirani2005cluster}: we generated $n = 50$  equally spaced points along the diagonal of a 3-dimensional cube with side length $[-5, 5]$, and added uncorrelated Gaussian noise ($\mu = 0$ and $\sigma_i = 0.1$) to each data point. We then copy these data points $k^*=3$ (scenario 3) or $k^*=7$ (scenario 4) times and place them along the same line separated by a distance of $15$. Similarly to above, the total sample size of scenario 3 and 4 is 150 and 350, respectivey. We provide code to fully reproduce our simulation results in the Online Supplementary Material. Columns three and four in Figure \ref{fig:simsetup} illustrate these elongated clusters in the first two dimensions, respectively.

\subsection{Comparison plan}

The main contribution of this paper is to improve existing stability-based methods. Our goal of the numerical experiments is therefore to see whether normalized clustering instability outperforms the unnormalized (standard) clustering instability in detecting $k^*$.

However, to also learn about the relative merits of stability-based methods, we compare their performance to the performance of popular distance-based methods for $k^*$-selection. Note that these methods imply different definitions of a 'good' clustering (see introduction). Thus, strictly speaking the different methods solve different problems. Nonetheless, in practice, all of these methods are applied for the same purpose. In this context, the various methods can be understood as different heuristics solutions to a given problem (here the 4 scenarios described in Section \ref{sim_data}).

We consider the following four distance-based methods: the Gap Statistic \citep{tibshirani2001estimating}, the Jump statistic \citep{sugarfinding}, the the  Silhouette statistic \citep{rousseeuw1987silhouettes}, the Slope statistic \citep{fujita2014non}, and a Gaussian mixture model. The Gap statistic simulates uniform data of the same dimensionality as the original data and then compares the gap between the logarithm of the within-cluster dissimilarity $W(k)$ for the simulated and original data. It selects the $k$ for which this gap is largest. The Jump statistic computes the differences of the within-cluster distortion at $k$ and $k-1$ (after transformation via a negative power) to select $k$ that produced the largest differences in distortions. The Slope statistic is based on the Silhouette statistic $Si()$, and selects $k$ to maximize $[Si(k) - Si(k-1) ] Si(k)^v$, where $v$ is a tuning parameter. Finally, the Gaussian mixture model selects $k$ as the number of components in the mixture model yielding the lowest Bayesian Information Criterion (BIC) \citep{schwarz1978estimating}.

\subsection{Results}\label{sim_results}

We evaluated the $k$-selection methods using the k-means clustering algorithm \citep{hartigan1975clustering}. The k-means algorithm was restarted 10 times with random starting centroids in order to avoid local minima. \cite{dick2014many} showed that 10 restarts for k-means were sufficient for two clustering problems that match the problems considered here in difficulty. For all methods, we considered the sequence $k = \{2, 3, \dots, 50\}$. For the stability-based methods we use 100 bootstrap comparisons (see Algorithm~\ref{algorithm1} and ~\ref{algorithm2}). To maximize comparability, we evaluated used the same random seed for all instability-based methods (within the same iteration).

\begin{table}[H]
	\centering
	\scalebox{0.75}{
		\begin{tabular}{lccccccccccccccccccc}
			\\
			&\multicolumn{19}{c}{3 circular clusters, 2 dimensions} \\
			& 2 & \textbf{3} & 4 & 5 & 6 & 7 & 8 & 9 & 10 & 11 & 12 & 13 & 14 & 15 & 16 & 17& 18& 19& 20+  \\[5pt]
			Model-based & 0 & \textbf{68} & 0 & 0 & 0 & 0 & 0 & 0 & 0 & 0 & 0 & 0 & 0 & 0 & 0 & 0 & 0 & 0 & 32 \\
			Model-based (N) & 0 & \textbf{100} & 0 & 0 & 0 & 0 & 0 & 0 & 0 & 0 & 0 & 0 & 0 & 0 & 0 & 0 & 0 & 0 & 0\\
			Model-free & 0 & \textbf{43} & 0 & 0 & 0 & 0 & 0 & 0 & 0 & 0 & 0 & 0 & 0 & 0 & 0 & 0 & 0 & 0 & 57\\
			Model-free (N) &0  & \textbf{100} & 0 & 0 & 0 & 0 & 0 & 0 & 0 & 0 & 0 & 0 & 0 & 0 & 0 & 0 & 0 & 0 & 0\\
			Gap Statistic & 0 & \textbf{100} & 0 & 0 & 0 & 0 & 0 & 0 & 0 & 0 & 0 & 0 & 0 & 0 & 0 & 0 & 0 & 0 & 0\\
			Jump Statistic & 0 & \textbf{0} & 0 & 0 & 0 & 0 & 0 & 0 & 0 & 0 & 0 & 0 & 0 & 0 & 0 & 0 & 0 & 0 & 100\\
			Slope Statistic & 0 & \textbf{96} & 4 & 0 & 1 & 0 & 0 & 0 & 0 & 0 & 0 & 0 & 0 & 0 & 0 & 0 & 0 & 0 & 0\\
			Gaussian Mixture & 0 & \textbf{100} & 0 & 0 & 0 & 0 & 0 & 0 & 0 & 0 & 0 & 0 & 0 & 0 & 0 & 0 & 0 & 0 & 0\\
			\\
			&\multicolumn{19}{c}{7 circular clusters, 2 dimensions} \\
			& 2 & 3 & 4 & 5 & 6 & \textbf{7} & 8 & 9 & 10 & 11 & 12 & 13 & 14 & 15 & 16 & 17& 18& 19& 20+  \\[5pt]
			Model-based         & 0 & 0 & 0 & 0 & 0 & \textbf{0} & 0 & 0 & 0 & 0 & 0 & 0 & 0 & 0 & 0 & 0 & 0 & 0 & 100 \\
			Model-based (N)    & 0 & 0 & 0 & 0 & 0 & \textbf{87} & 13 & 0 & 0 & 0 & 0 & 0 & 0 & 0 & 0 & 0 & 0 & 0 & 0\\
			Model-free      & 0 & 0 & 0 & 0 & 0 & \textbf{0} & 0 & 0 & 0 & 0 & 0 & 0 & 0 & 0 & 0 & 0 & 0 & 0 & 100\\
			Model-free (N) & 0 & 0 & 0 & 0 & 0 & \textbf{91} & 9 & 0 & 0 & 0 & 0 & 0 & 0 & 0 & 0 & 0 & 0 & 0 & 0\\
			Gap Statistic    & 0 & 0 & 0 & 0 & 0 & \textbf{100} & 0 & 0 & 0 & 0 & 0 & 0 & 0 & 0 & 0 & 0 & 0 & 0 & 0\\
			Jump Statistic & 0 & 0 & 0 & 0 & 0 & \textbf{8} & 0 & 0 & 0 & 0 & 0 & 0 & 0 & 0 & 0 & 0 & 0 & 0 & 92\\
			Slope Statistic & 0 & 0 & 0 & 0 & 0 & \textbf{31} & 17 & 18 & 18 & 9 & 5 & 1 & 0 & 0 & 1 & 0 & 0 & 0 & 0\\
			Gaussian Mixture & 0 & 0 & 0 & 0 & 0 & \textbf{99}& 1 & 0 & 0 & 0 & 0 & 0 & 0 & 0 & 0 & 0 & 0 & 0 & 0\\
			\\
			&\multicolumn{19}{c}{3 elongated clusters, 2 dimensions} \\
			& 2 & \textbf{3} & 4 & 5 & 6 & 7 & 8 & 9 & 10 & 11 & 12 & 13 & 14 & 15 & 16 & 17& 18& 19& 20+  \\[5pt]
			Model-based & 0 & \textbf{100} & 0 & 0 & 0 & 0 & 0 & 0 & 0 & 0 & 0 & 0 & 0 & 0 & 0 & 0 & 0 & 0 & 0 \\
			Model-based (N) & 0 & \textbf{100} & 0 & 0 & 0 & 0 & 0 & 0 & 0 & 0 & 0 & 0 & 0 & 0 & 0 & 0 & 0 & 0 & 0\\
			Model-free & 0 & \textbf{100} & 0 & 0 & 0 & 0 & 0 & 0 & 0 & 0 & 0 & 0 & 0 & 0 & 0 & 0 & 0 & 0 & 0\\
			Model-free (N) &0  & \textbf{100} & 0 & 0 & 0 & 0 & 0 & 0 & 0 & 0 & 0 & 0 & 0 & 0 & 0 & 0 & 0 & 0 & 0\\
			Gap Statistic & 0 & \textbf{0} & 0 & 0 & 0 & 0 & 0 & 0 & 0 & 0 & 0 & 0 & 0 & 0 & 0 & 0 & 0 & 0 & 100\\
			Jump Statistic & 0 & \textbf{0} & 0 & 0 & 0 & 0 & 0 & 0 & 0 & 0 & 0 & 0 & 0 & 0 & 0 & 0 & 0 & 0 & 100\\
			Slope Statistic & 0 & \textbf{100} & 0 & 0 & 0 & 0 & 0 & 0 & 0 & 0 & 0 & 0 & 0 & 0 & 0 & 0 & 0 & 0 & 0\\
			Gaussian Mixture & 0 & \textbf{99} & 0 & 0 & 1 & 0 & 0 & 0 & 0 & 0 & 0 & 0 & 0 & 0 & 0 & 0 & 0 & 0 & 0\\
			\\
			&\multicolumn{19}{c}{7 elongated clusters, 2 dimensions} \\
			& 2 & 3 & 4 & 5 & 6 & \textbf{7} & 8 & 9 & 10 & 11 & 12 & 13 & 14 & 15 & 16 & 17& 18& 19& 20+  \\[5pt]
			Model-based          & 0 & 0 & 0 & 0 & 0 & \textbf{0} & 0 & 0 & 0 & 0 & 0 & 0 & 0 & 0 & 0 & 0 & 0 & 0 & 100 \\
			Model-based (N)     & 24 & 0 & 0 & 0 & 0 & \textbf{38} & 38 & 1 & 1 & 0 & 0 & 0 & 0 & 0 & 0 & 0 & 0 & 0 & 0\\
			Model-free      & 0 & 0 & 0 & 0 & 0 & \textbf{0} & 0 & 0 & 0 & 0 & 0 & 0 & 0 & 0 & 0 & 0 & 0 & 0 & 100\\
			Model-free (N) & 24 & 0 & 0 & 0 & 0 & \textbf{47} & 29 & 0 & 0 & 0 & 0 & 0 & 0 & 0 & 0 & 0 & 0 & 0 & 0\\
			Gap Statistic    & 0 & 0 & 0 & 0 & 0 & \textbf{0} & 0 & 0 & 0 & 0 & 0 & 0 & 0 & 0 & 0 & 0 & 0 & 0 & 100\\
			Jump Statistic & 0 & 0 & 0 & 0 & 0 & \textbf{0} & 0 & 0 & 0 & 0 & 0 & 0 & 0 & 0 & 0 & 0 & 0 & 0 & 100\\
			Slope Statistic & 0 & 0 & 0 & 0 & 0 & \textbf{68} & 7 & 10 & 6 & 2 & 4 & 3 & 0 & 0 & 0 & 0 & 0 & 0 & 0\\
			Gaussian Mixture & 0 & 0 & 0 & 0 & 0 & \textbf{100} & 0 & 0 & 0 & 0 & 0 & 0 & 0 & 0 & 0 & 0 & 0 & 0 & 0\\
			\\
		\end{tabular}%
	}
	\caption{Estimated number of clusters in four different scenarios for 100 iterations.}
	\label{tb:results}
\end{table}

Table \ref{tb:results} shows the estimated $\hat{k}$ over 100 iterations for each of the four scenarios and eight methods. Estimated $\hat{k} \geq 20$ are collapsed in the category '20+'. We first focus on the results of the instability-based methods. For the first scenario with $k^*=3$ circular clusters, the unnormalized instability-based methods perform poorly, with about half of the estimates being correct, and the other half being in the category '20+'. This poor performance is due to the unfavorable behavior illustrated in Figure \ref{fig:instabcurve} and Figure \ref{fig:illustration}. The normalized instability methods, however, do not suffer from this problem and accordingly show high performance. The pattern of results in the scenario with $k^* = 7$ is similar, but more pronounced. With the clustering problem being more difficult, unnormalized stability methods fail to identify $k^*$ in every iteration, whereas the normalized stability methods still successfully identify $k^*$ in the vast majority of cases. In scenario three with $k^*=3$ elongated clusters all instability-based methods show maximum performance. Here the tail of the unnormalized instability paths decays slowly enough preventing the trivial decrease instability to undercut the local minimum at $k = 3$.  In scenario four with $k^*=7$ elongated clusters, the performance of the unnormalized methods drops to zero, whereas normalized instability methods are still able to identify $k^*$ in a considerable amount of cases. Overall the results show that the normalized instability-methods perform better than the unnormalized ones.

We now turn to the performance of distance-based methods. The clear winner among this class of methods is the Gaussian mixture, which performs performs extremely well in all scenarios. Next, the Slope statistic performs reasonably well, however, the performance is much lower for $k^*=7$ than for  $k^*=3$. The Gap statistic shows maximal performance for the circular clusterings, but drops to zero in for the elongated clusters. Finally, the Jump statistic fails entirely in all scenarios. The reason for the bad performance of the Jump statistic is that its variance increases with increasing $k$. See Appendix \ref{app_jump} for a detailed illustration of this problem. 

Comparing stability- and distance-based methods, we find that stability-based methods perform rather well. When using our proposed normalization, stability-based methods outperform every distance-based method, except for the Gaussian mixture methods. However, when using the unnormalized methods, it is almost always better to use any of the distance-based methods. For additional comparisons between the methods, consult  Appendix \ref{app_additional} where we study small variations of scenario one and two including additional noise dimensions. 

\begin{figure}
	\centering
	\includegraphics[width=.9\linewidth]{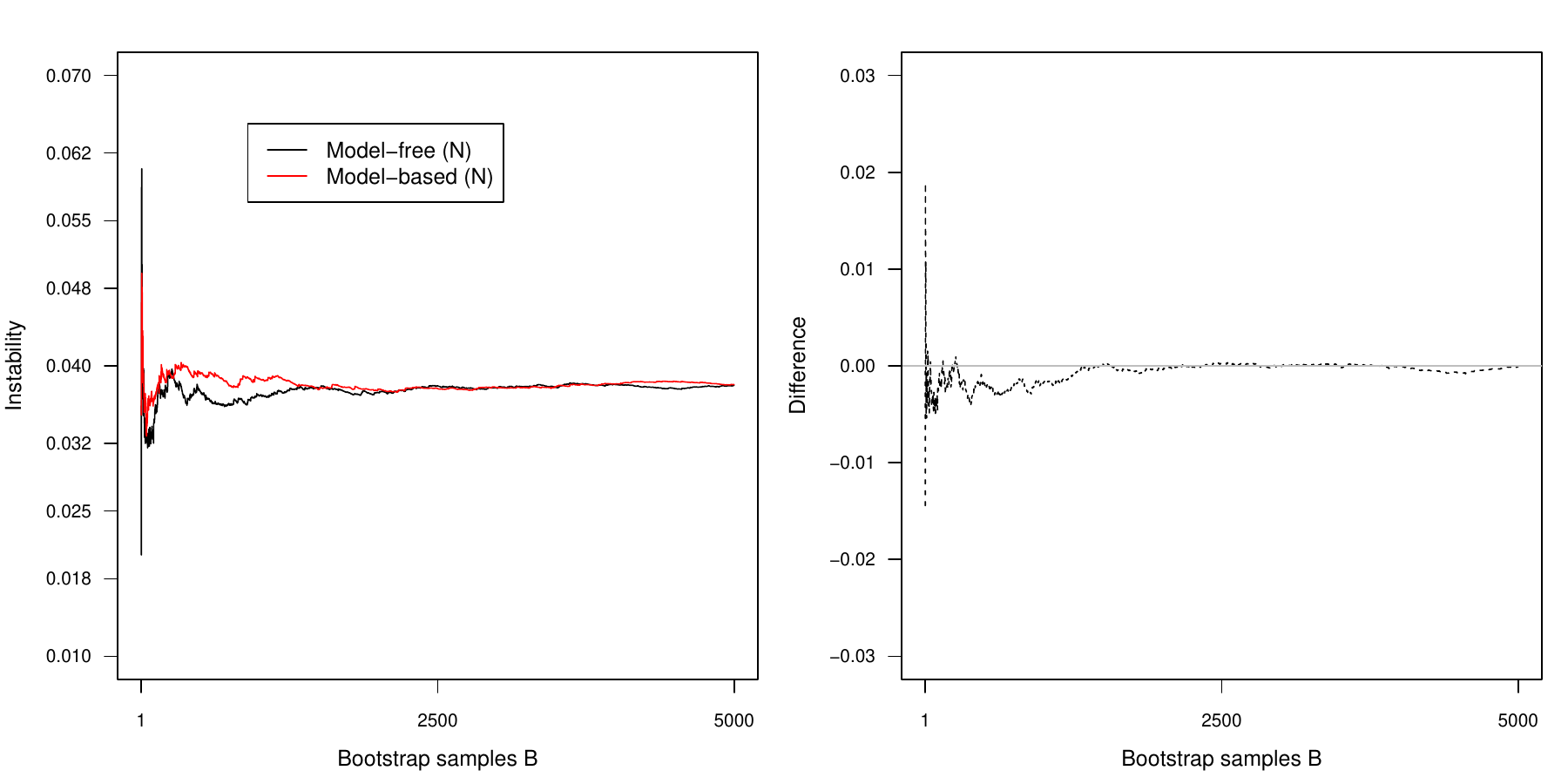}
	\caption{Left: The average instability for fixed $k=3$ up to bootstrap sample $b$ for both the model-based (red) and model-free (black) instability approach. The data is the one from in Figure \ref{fig:illustration}. Right: The difference between the two functions.}
	\label{fig:convergence}
\end{figure}

Another noteworthy finding of our analysis is the near-equivalent performance of the model-based and the model-free instability approaches (see Table \ref{tb:results}). To analyze whether the two methods converge for large $B$, we ran both methods using the scenario of Figure \ref{fig:illustration}) over a increasing number of $B \in \{1, 2, \dots, 5000\}$ bootstrap comparisons. Figure \ref{fig:convergence} shows that although both methods seem to stabilize in a small region around $.038$ they still show considerable variance even with 5000 bootstrap samples. It is thus unclear whether the two methods converge, however, they may converge for larger $B$.

Furthermore, we evaluated the correlation between the instability paths of both approaches for the simulation reported in Table \ref{tb:results}. They are between $.98$ and $1$, suggesting that the two methods show very similar performance.

\section{Conclusions}

We have proposed a normalization for cluster-instability methods for the estimation of $k^*$ and shown how it improves existing methods by controlling for the influence of chance on cluster instability and enabling the identification of $k^*$ for the entire range of $k$. Moreover, we have shown that instability-based methods, especially when using the proposed normalization, can outperform established distance-based $k$-selection methods. Finally, we have compared model-based and model-free variants of the instability-based method and found them to be similar, but not identical. 

These results speak to the usefulness of cluster instability as an approach for estimating the number of clusters in a dataset. They also lead to two important questions for future research. First, given the divergence of the model-based and model-free approaches, future research should study in closer detail the relative performance of the two across different situations. Second, and more importantly, future research should study potentially more appropriate corrections for the influence of chance on cluster instability. As $d^r(\psi_a, \psi_b)$ depends on $M$, $d^r$ it will also be susceptible to factors influencing $M$ such as the data generating mechanism $\mathcal{P}$ and the clustering algorithm. While our numerical experiments demonstrate the usefulness of $d^n$ as defined here, we see potential for more complex alternative corrections.

\begin{acknowledgements}
We would like to thank Junhui Wang for helpful answers to several questions about his papers on clustering instability. JMBH was supported by the European Research Council Consolidator Grant no. 647209.
\end{acknowledgements}

\bibliographystyle{spbasic}      
\bibliography{stab_paper}   

\begin{thebibliography}{16}
\providecommand{\natexlab}[1]{#1}
\providecommand{\url}[1]{{#1}}
\providecommand{\urlprefix}{URL }
\expandafter\ifx\csname urlstyle\endcsname\relax
  \providecommand{\doi}[1]{DOI~\discretionary{}{}{}#1}\else
  \providecommand{\doi}{DOI~\discretionary{}{}{}\begingroup
  \urlstyle{rm}\Url}\fi
\providecommand{\eprint}[2][]{\url{#2}}

\bibitem[{Ben-Hur et~al(2001)Ben-Hur, Elisseeff, and Guyon}]{ben2001stability}
Ben-Hur A, Elisseeff A, Guyon I (2001) A stability based method for discovering
  structure in clustered data. In: Pacific symposium on biocomputing, vol~7, pp
  6--17

\bibitem[{Bengio et~al(2003)Bengio, Vincent, Paiement, Delalleau, Ouimet, and
  Le~Roux}]{bengio2003spectral}
Bengio Y, Vincent P, Paiement JF, Delalleau O, Ouimet M, Le~Roux N (2003)
  Spectral clustering and kernel PCA are learning eigenfunctions, vol 1239.
  Citeseer

\bibitem[{Dick et~al(2014)Dick, Wong, and Dann}]{dick2014many}
Dick T, Wong E, Dann C (2014) How many random restarts are enough? Google
  Scholar

\bibitem[{Fang and Wang(2012)}]{FangWang2012}
Fang Y, Wang J (2012) Selection of the number of clusters via the bootstrap
  method. Computational Statistics \& Data Analysis 56(3):468--477

\bibitem[{Friedman et~al(2001)Friedman, Hastie, and
  Tibshirani}]{friedman2001elements}
Friedman J, Hastie T, Tibshirani R (2001) The elements of statistical learning,
  vol~1. Springer series in statistics Springer, Berlin

\bibitem[{Fujita et~al(2014)Fujita, Takahashi, and Patriota}]{fujita2014non}
Fujita A, Takahashi DY, Patriota AG (2014) A non-parametric method to estimate
  the number of clusters. Computational Statistics \& Data Analysis 73:27--39

\bibitem[{Hartigan(1975)}]{hartigan1975clustering}
Hartigan JA (1975) Clustering algorithms

\bibitem[{Hennig(2007)}]{hennig2007cluster}
Hennig C (2007) Cluster-wise assessment of cluster stability. Computational
  Statistics \& Data Analysis 52(1):258--271

\bibitem[{Hennig(2015)}]{hennig2015true}
Hennig C (2015) What are the true clusters? Pattern Recognition Letters
  64:53--62

\bibitem[{Ng et~al(2002)Ng, Jordan, Weiss et~al}]{ng2002spectral}
Ng AY, Jordan MI, Weiss Y, et~al (2002) On spectral clustering: Analysis and an
  algorithm. Advances in neural information processing systems 2:849--856

\bibitem[{Rousseeuw(1987)}]{rousseeuw1987silhouettes}
Rousseeuw PJ (1987) Silhouettes: a graphical aid to the interpretation and
  validation of cluster analysis. Journal of computational and applied
  mathematics 20:53--65

\bibitem[{Schwarz et~al(1978)}]{schwarz1978estimating}
Schwarz G, et~al (1978) Estimating the dimension of a model. The annals of
  statistics 6(2):461--464

\bibitem[{Sugar and James(2011)}]{sugarfinding}
Sugar CA, James GM (2011) Finding the number of clusters in a dataset. Journal
  of the American Statistical Association

\bibitem[{Tibshirani and Walther(2005)}]{tibshirani2005cluster}
Tibshirani R, Walther G (2005) Cluster validation by prediction strength.
  Journal of Computational and Graphical Statistics 14(3):511--528

\bibitem[{Tibshirani et~al(2001)Tibshirani, Walther, and
  Hastie}]{tibshirani2001estimating}
Tibshirani R, Walther G, Hastie T (2001) Estimating the number of clusters in a
  data set via the gap statistic. Journal of the Royal Statistical Society:
  Series B (Statistical Methodology) 63(2):411--423

\bibitem[{Wang(2010)}]{wang2010consistent}
Wang J (2010) Consistent selection of the number of clusters via
  crossvalidation. Biometrika 97(4):893--904

\end{thebibliography}

\appendix

\section{Additional Numerical Experiments}\label{app_additional}

We show the results of two additional scenarios that are adapted from scenarios one and two in Section \ref{sim_results} by adding eight dimensions of uncorrelated Gaussian noise with standard deviations $\sigma = .15$ in scenario one and $\sigma = .04$ in scenario two.

\begin{table}[H]
	\label{tb:results_app}
	\centering
	\scalebox{0.75}{
		\begin{tabular}{lccccccccccccccccccc}
			\\
			&\multicolumn{19}{c}{3 circular clusters, 10 dimensions} \\
			\\
			& 2 & \textbf{3} & 4 & 5 & 6 & 7 & 8 & 9 & 10 & 11 & 12 & 13 & 14 & 15 & 16 & 17& 18& 19& 20+  \\[5pt]
			Model-based & 0 & \textbf{49} & 0 & 0 & 0 & 0 & 0 & 0 & 0 & 0 & 0 & 0 & 0 & 0 & 0 & 0 & 0 & 0 & 51 \\
			Model-based (N) & 0 & \textbf{100} & 0 & 0 & 0 & 0 & 0 & 0 & 0 & 0 & 0 & 0 & 0 & 0 & 0 & 0 & 0 & 0 & 0\\
			Model-free & 0 & \textbf{4} & 0 & 0 & 0 & 0 & 0 & 0 & 0 & 0 & 0 & 0 & 0 & 0 & 0 & 0 & 0 & 0 & 96\\
			Model-free (N) & 0 & \textbf{100} & 0 & 0 & 0 & 0 & 0 & 0 & 0 & 0 & 0 & 0 & 0 & 0 & 0 & 0 & 0 & 0 & 0\\
			Gap Statistic & 0 & \textbf{100} & 0 & 0 & 0 & 0 & 0 & 0 & 0 & 0 & 0 & 0 & 0 & 0 & 0 & 0 & 0 & 0 & 0\\
			Jump Statistic & 0 & \textbf{0} & 0 & 0 & 0 & 0 & 0 & 0 & 0 & 0 & 0 & 0 & 0 & 0 & 0 & 0 & 0 & 0 & 100\\
			Slope Statistic & 0 & \textbf{97} & 3 & 0 & 0 & 0 & 0 & 0 & 0 & 0 & 0 & 0 & 0 & 0 & 0 & 0 & 0 & 0 & 0\\
			Gaussian Mixture & 0 & \textbf{100} & 0 & 0 & 0 & 0 & 0 & 0 & 0 & 0 & 0 & 0 & 0 & 0 & 0 & 0 & 0 & 0 & 0\\
			\\
			&\multicolumn{19}{c}{7 circular clusters, 10 dimensions} \\
			\\
			& 2 & 3 & 4 & 5 & 6 & \textbf{7} & 8 & 9 & 10 & 11 & 12 & 13 & 14 & 15 & 16 & 17& 18& 19& 20+  \\[5pt]
			Model-based          & 0 & 0 & 0 & 0 & 0 & \textbf{0} & 0 & 0 & 0 & 0 & 0 & 0 & 0 & 0 & 0 & 0 & 0 & 0 & 100 \\
			Model-based (N)     & 0 & 0 & 0 & 0 & 0 & \textbf{68} & 32 & 0 & 0 & 0 & 0 & 0 & 0 & 0 & 0 & 0 & 0 & 0 & 0\\
			Model-free      & 0 & 0 & 0 & 0 & 0 & \textbf{0} & 0 & 0 & 0 & 0 & 0 & 0 & 0 & 0 & 0 & 0 & 0 & 0 & 100\\
			Model-free(N) & 0 & 0 & 0 & 0 & 0 & \textbf{83} & 17 & 0 & 0 & 0 & 0 & 0 & 0 & 0 & 0 & 0 & 0 & 0 & 0\\
			Gap Statistic    & 0 & 0 & 0 & 0 & 0 & \textbf{100} & 0 & 0 & 0 & 0 & 0 & 0 & 0 & 0 & 0 & 0 & 0 & 0 & 0\\
			Jump Statistic & 0 & 0 & 0 & 0 & 0 & \textbf{0} & 0 & 0 & 0 & 0 & 0 & 0 & 0 & 0 & 0 & 0 & 0 & 0 & 100\\
			Slope Statistic & 0 & 9 & 91 & 0 & 0 & \textbf{0} & 0 & 0 & 0 & 0 & 0 & 0 & 0 & 0 & 0 & 0 & 0 & 0 & 0\\
			Gaussian Mixture & 0 & 0 & 0 & 0 & 0 & \textbf{100} & 0 & 0 & 0 & 0 & 0 & 0 & 0 & 0 & 0 & 0 & 0 & 0 & 0\\
			\\
		\end{tabular}%
	}
	\caption{Estimated number of clusters in different scenarios}
	
\end{table}

The performance is qualitativly similar to the performance reported in the main text.  However, performance dropped for all methods as a result of the added noise, which rendered the clustering problem more difficult.

\section{Path of Jump Statistic}\label{app_jump}

\begin{figure}
	\centering
	\includegraphics[width=1\linewidth]{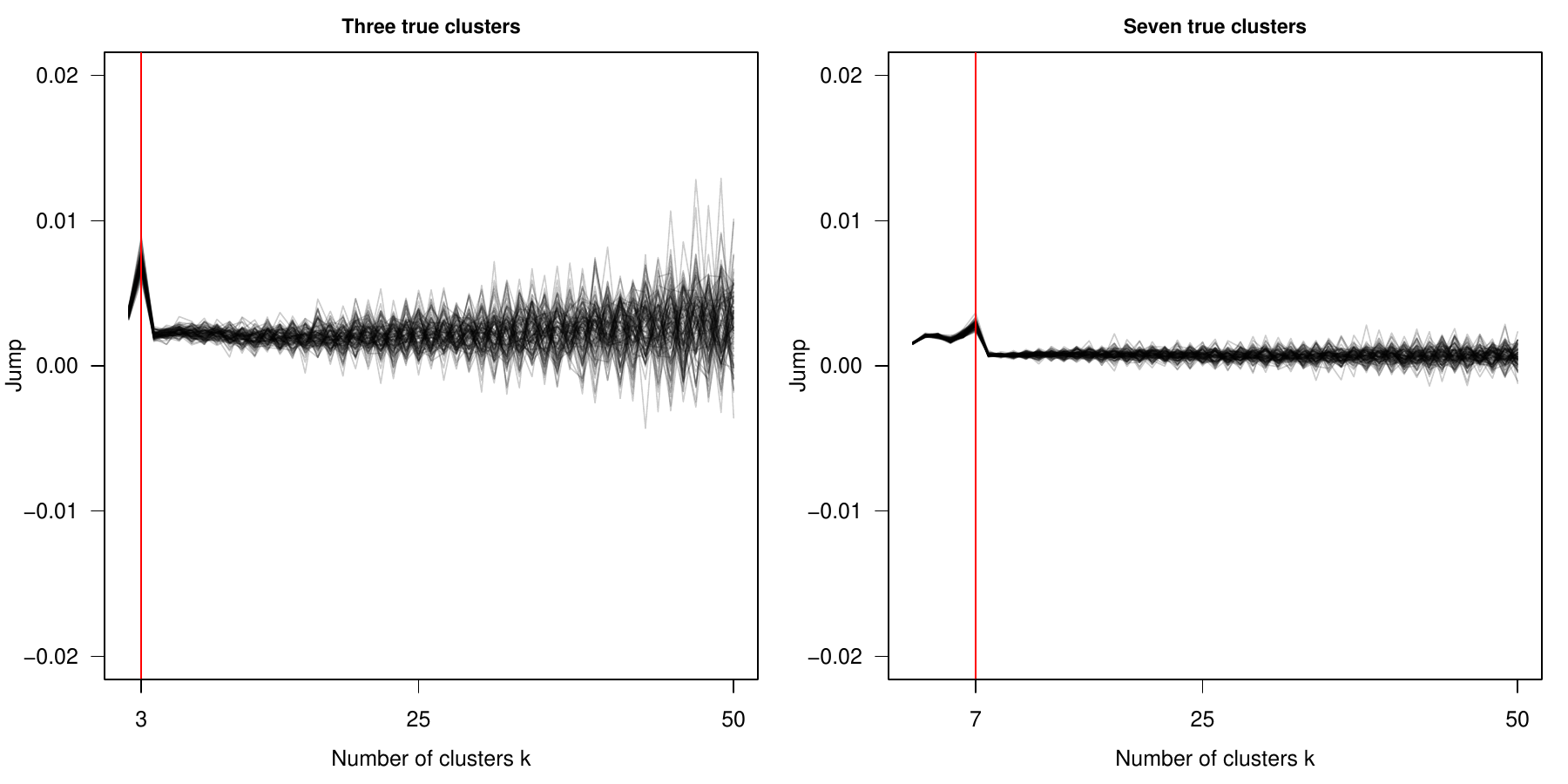}
	\caption{The jump-path along the sequence of considered $k$ for each of the 100 simulation iterations. Left: for the scenario with 3 true clusters in 2 dimensions. Right: for the scenario with 7 true clusters in 2 dimensions.}
	\label{fig:jump}
\end{figure}

One reason for the bad performance of the Jump statistic is that the variance of the jump size increasing as $k$ increases. We illustrate this problematic behavior of the jump statistic in Figure \ref{fig:jump} using 100 iterations of scenario one (three circular clusters) and two (seven circular clusters) from the main text.

The figure plots are the paths Jump statistic for each of the 100 iterations across $k \in \{2, 3, \dots, 10\}$. We see that the variance of the Jump statistic paths clearly increases for larger $k$. This implies that   similarly to the unnormalized stability-based methods, the Jump statistic can only identify $k^*$, when the range of possible $k$ is restricted to a small range around the true $k^*$.

\end{document}